# Lasso type classifiers with a reject option

Marten Wegkamp[*]

*Department of Statistics, Florida State University, Tallahassee, FL 32306-4330*
*e-mail:* `wegkamp@stat.fsu.edu`

**Abstract:** We consider the problem of binary classification where one can, for a particular cost, choose not to classify an observation. We present a simple proof for the oracle inequality for the excess risk of structural risk minimizers using a lasso type penalty.

Received May 2007.

## 1. Introduction

This paper discusses structural risk minimization in the setting of classification with a reject option. Binary classification is about classifying observations that take values in an arbitrary feature space $\mathcal{X}$ into one of two classes, labelled $-1$ or $+1$. A discriminant function $f : \mathcal{X} \to \mathbb{R}$ yields a classifier $\text{sgn}(f(x)) \in \{-1, +1\}$ that represents our guess of the label $Y$ of a future observation $X$ and we err if the margin $y \cdot f(x) < 0$. Since observations $x$ for which the conditional probability

$$\eta(x) = \mathbb{P}\{Y = +1 | X = x\} \tag{1}$$

is close to $1/2$ are difficult to classify, we introduce a reject option for classifiers, by allowing for a third decision, ⓡ (reject), expressing doubt.

We built in the reject option by using a threshold value $0 \leq \tau < 1$ as follows. Given a discriminant function $f : \mathcal{X} \to \mathbb{R}$, we report $\text{sgn}(f(x)) \in \{-1, 1\}$ if $|f(x)| > \tau$, but we withhold decision if $|f(x)| \leq \tau$ and report ⓡ. We assume that the cost of making a wrong decision is 1 and the cost of utilizing the reject option is $d > 0$. The appropriate risk function is then

$$\mathbb{E}\left[\ell(Yf(X))\right] = \mathbb{P}\{Yf(X) < -\tau\} + d\mathbb{P}\{|Yf(X)| \leq \tau\} \tag{2}$$

[*]Research is supported in part by NSF grant DMS 0706829

for the discontinuous loss

$$\ell(z) = \begin{cases} 1 & \text{if } z < -\tau, \\ d & \text{if } |z| \leq \tau, \\ 0 & \text{otherwise.} \end{cases} \tag{3}$$

Since we never reject if $d > 1/2$, see [11], we restrict ourselves to the cases $0 \leq d \leq 1/2$. The generalized Bayes discriminant function, minimizing (2), is then

$$f_0(x) = \begin{cases} -1 & \text{if } \eta(x) < d \\ 0 & \text{if } d \leq \eta(x) \leq 1 - d \\ +1 & \text{if } \eta(x) > 1 - d \end{cases} \tag{4}$$

with risk

$$\mathbb{E}\left[\min\{\eta(X), 1 - \eta(X), d\}\right],$$

see [9, 13]. The case $(\tau, d) = (0, 1/2)$ reduces to the classical situation without the reject option. We can view $d$ as an upper bound on the conditional probability of misclassification (given $X$) that is considered tolerable.

The estimators

$$\mathsf{f}_\lambda(x) = \sum_{i=1}^M \lambda_i f_i(x), \quad \lambda \in \mathbb{R}^M,$$

of $f_0(x)$ that we study in this paper are linear combinations of base functions $f_j$ from a dictionary $F_M = \{f_1, \ldots, f_M\}$. We suggest regularized empirical risk minimization based using convex surrogate loss functions $\phi$ and a penalty term $p(\lambda) = 2r_n|\lambda|_1$ that is proportional to the $\ell_1$-norm $|\lambda|_1$ of the parameter $\lambda$. The regularized empirical risk

$$\frac{1}{n}\sum_{i=1}^n \phi(Y_i \mathsf{f}_\lambda(X_i)) + p(\lambda) \tag{5}$$

is then convex in $\lambda$ and its minimization can be solved by a (tractable) convex program.

The organization of the paper is as follows. Section 2 presents a general bound on the excess risk of minimizers $\widehat{\lambda}$ of the penalized empirical risk (5). We define an oracle target $\lambda^*$, that provides an ideal approximation $\mathsf{f}_{\lambda^*}$ of $f_0$ with possibly many fewer elements $f_i$ of the dictionary $F_M$, and show under mild assumptions that this oracle target can be recovered by minimization of (5), even if $M$ is larger than $n$. We advance the use of a novel type of oracle inequality, explored in [8, 6], where the aim is to show that the sum of the excess risk and the penalty term $p(\widehat{\lambda} - \lambda^*)$ achieves the optimal balance between the excess risk and a regularization term. This allows us to determine that the oracle can be recovered and gives us information about the $\ell_1$-distance between $\widehat{\lambda}$ and the oracle vector $\lambda^*$. This extends the work of [4, 5, 6, 7] on lasso-type estimators in regression and density estimation problems to empirical risk minimization of the general criterion (5) in the context of classification with a reject option. We take a different approach than the recent technical report [17]. In particular, we use the concept of mutual coherence, used in [4, 5, 6, 7], which is weaker than the corresponding requirement in [17] and give a different, simple proof of the main oracle inequality. We demonstrate that the choice of the the tuning parameter $r_n$ in the penalty $p(\lambda) = 2r_n|\lambda|_1$ is crucial. We prove that the oracle inequality holds on an event where $r_n$ exceeds a certain random quantity $\widehat{r}$. Then we show that $\widehat{r}$ is highly concentrated around its mean using McDiarmid's concentration inequality and provide an upper bound for $\mathbb{E}[\widehat{r}]$.

Section 3 applies the results of Section 2 to the specific generalized hinge loss function $\phi_d$ introduced in [1], extending the work [14] to classification with a reject option. This loss is convex, so that the minimization of (5) is computationally feasible, and at the same time classification calibrated, as the minimizer of $\mathbb{E}[\phi_d(Yf(X))]$ is the Bayes discriminant $f_0$, our parameter of interest.

Finally, the proofs are collected in Section 4.

## 2. Oracle inequalities for the excess risk

### 2.1. Preliminaries

The data $(X_1, Y_1), \ldots, (X_n, Y_n)$ consist of independent copies of $(X, Y)$ where $X$ takes values in an arbitrary measurable space $\mathcal{X}$ and $Y \in \{-1, +1\}$. Let $F_M = \{f_1, \ldots, f_M\}$ be a finite set of functions (dictionary) with $\|f_j\|_\infty \leq C_F$ and we consider discriminant functions

$$\mathsf{f}_\lambda(x) = \sum_{j=1}^M \lambda_j f_j(x), \quad \lambda \in \mathbb{R}^M.$$

We consider a loss function $\phi: \mathbb{R} \to [0, \infty)$ that is Lipschitz,

$$|\phi(y) - \phi(y')| \leq C_\phi |y - y'|$$

with $C_\phi < \infty$ and based on this loss function, we define the risk functions

$$R_\phi(\lambda) = \mathbb{E}\left[\phi(Y\mathsf{f}_\lambda(X))\right] \quad \text{and} \quad \widehat{R}_\phi(\lambda) = \frac{1}{n}\sum_{i=1}^n \phi(Y_i\mathsf{f}_\lambda(X_i)).$$

We assume that $f_0$ defined in (4) minimizes the risk $\mathbb{E}[\phi(Yf(X))]$ over all measurable $f: \mathcal{X} \to \mathbb{R}$, and we denote its risk by $R_0$, that is,

$$R_0 = \inf_f \mathbb{E}\left[\phi(Yf(X))\right]$$

We measure the performance of our estimators in terms of the excess risk

$$\Delta_\phi(\lambda) = R_\phi(\lambda) - R_0.$$

Based on the penalty

$$p(\lambda) = 2r_n |\lambda|_1 = 2r_n \sum_{i=1}^{M} |\lambda_i|$$

with $r_n$ specified later in Section 2.4, the penalized empirical risk minimizer $\widehat{\lambda}$ satisfies

$$\widehat{R}_\phi(\widehat{\lambda}) + p(\widehat{\lambda}) \leq \widehat{R}_\phi(\lambda) + p(\lambda) \quad \text{for all } \lambda \in \mathbb{R}^M. \tag{6}$$

In particular, (6) ensures that for $\lambda_0 = (0, \ldots, 0)$,

$$p(\widehat{\lambda}) \leq \widehat{R}_\phi(\widehat{\lambda}) + p(\widehat{\lambda}) \leq \widehat{R}_\phi(\lambda_0) + p(\lambda_0) = \phi(0)$$

which in turn implies $|\widehat{\lambda}|_1 \leq \phi(0)/(2r_n)$. This means that we effectively minimize the penalized empirical risk $\widehat{R}_\phi(\lambda) + p(\lambda)$ over $\lambda$ in the set

$$\Lambda_n = \left\{ \lambda \in \mathbb{R}^M : |\lambda|_1 \leq \phi(0)/(2r_n) \right\}.$$

## 2.2. Assumptions

We impose two conditions. Given some finite measure $\mu$ on $\mathcal{X}$, set

$$<f, g> = \int f(x) g(x)\, \mu(dx) \quad \text{and} \quad \|f\|^2 = \int f^2(x)\, \mu(dx).$$

The first condition imposes a link between the distance $\|\mathsf{f}_\lambda - f_0\|$ and excess risk $\Delta_\phi(\lambda)$:

**Condition 1.** *There exist $C_{\Delta,\mu} < \infty$ and $0 \leq \beta < 1$ such that, for all $\lambda \in \Lambda_n$,*

$$\|\mathsf{f}_\lambda - f_0\| \leq C_{\Delta,\mu} \Delta_\phi^\beta(\lambda). \tag{7}$$

In regression and density estimation problems as considered in [4, 5, 6, 7], this condition trivially holds with $\beta = 1/2$ and $C_{\Delta,\mu} = 1$. This relation is more delicate to establish in classification problems. It depends on the behavior of the conditional probability $\eta(X)$ near $d$ and $1 - d$, see Section 3 below.

Our goal is to estimate $f_0$ via linear combinations $\mathsf{f}_\lambda(x)$ and to evaluate performance in terms of the excess risk $\Delta_\phi(\lambda)$. For any $I = \{i_1, \ldots, i_m\} \subseteq \{1, \ldots, M\}$, we define the approximating parameter space

$$\Lambda(I) = \left\{ \lambda \in \mathbb{R}^M : \lambda_i = 0 \text{ for all } i \notin I \right\}$$

and let $\widehat{\lambda}_I$ minimize $\widehat{R}_\phi(\lambda)$ over $\Lambda(I)$. An oracle that knows $f_0$ would be able to tell us in advance which approximating space $\Lambda(I)$ yields the smallest excess risk $\Delta_\phi(\widehat{\lambda}_I)$. However, $f_0$ is unknown so the best we can do is to mimic the behavior of the oracle. General theory for empirical risk minimization in the classification context [2, 3, 11] indicates that

$$\Delta_\phi(\widehat{\lambda}_I) \lesssim \inf_{\lambda \in \Lambda(I)} \Delta_\phi(\lambda) + \left(\frac{|I|}{n}\right)^\rho,$$

where $|I|$ denotes the cardinality of the set $I$ and the symbol $\lesssim$ means that the inequality holds up to known multiplicative constants. Various choices are possible for the parameter $\rho$ depending on the margin exponent $\alpha \geq 0$ defined in Section 3. Our target of interest, the oracle vector $\lambda^* \in \Lambda_n$, depends on $\beta$. Formally, we define it as follows:

**Definition.** Let $c_\mu = \min_{1 \leq i \leq M} \|f_i\|$ and let $\lambda^*$ be the minimizer of

$$3\Delta_\phi(\lambda) + 2\left(\frac{8C_{\Delta,\mu}}{c_\mu}\right)^{\frac{1}{1-\beta}} \left(r_n^2 |\lambda|_0\right)^{\frac{1}{2-2\beta}}, \tag{8}$$

over $\lambda \in \Lambda_n$, where $|\lambda|_0 = \sum_{i=1}^M |\lambda_i|$ is the number of non-zero coefficients of the vector $\lambda$.

Thus $\lambda^*$ balances the approximation error, as measured by the excess risk $\Delta_\phi(\lambda)$, and the complexity of the parameter set $\Lambda(I)$ to which $\lambda^*$ belongs to, as measured by the regularization term $(r_n^2 |\lambda|_0)^{1/(2-2\beta)}$. The constants 3 and $2(8C_{\Delta,\mu})^{1/(1-\beta)}$ can be changed: A decrease in the former will lead to a increase in the latter, and vice-versa. The constant $c_\mu$ can be avoided altogether if we take the penalty $p(\lambda) = 2r_n \sum_{i=1}^M \|f_i\| |\lambda_i|$, but in practice $\mu$, and consequently $\|f_i\|$, is unknown. Surely we could plug in estimates for $\|f_i\|$ as in [4, 5, 6, 17], but we chose to keep the exposition and proofs as simple as possible.

Let

$$I^* = \{i : \lambda_i^* \neq 0\}$$

be the collection of non-zero coefficients of $\lambda^*$,

$$|\lambda^*|_0 = \sum_{i=1}^M I_{\{\lambda_i^* \neq 0\}}$$

be the cardinality of $I^*$, and

$$\rho(i,j) = \frac{<f_i, f_j>}{\|f_i\| \cdot \|f_j\|}$$

be the correlation between $f_i$ and $f_j$. Our second assumption requires that

$$\rho^* = \max_{i \in I^*} \max_{j \neq i} |\rho(i,j)| \tag{9}$$

is small:

**Condition 2.** *Let $c_\mu = \min_{1 \leq j \leq M} \|f_j\|$ and assume that*

$$12\rho^*|\lambda^*|_0 \leq c_\mu. \tag{10}$$

This mainly states that the submatrix $(<f_i, f_j>)_{i,j \in I^*}$ is positive definite and that the correlations $\rho(i,j)$ between elements $f_i$, $i \in I^*$, of this submatrix and outside elements $f_j$, $j \notin I^*$, are relatively small. We refer to this assumption as the local mutual coherence assumption, see [4, 5, 6, 7].

### 2.3. Oracle inequality

Instrumental in our argument is the random quantity

$$\widehat{r} = \sup_{\lambda \in \Lambda_n} \frac{\left|(\widehat{R}_\phi - R_\phi)(\lambda) - (\widehat{R}_\phi - R_\phi)(\lambda^*)\right|}{|\lambda - \lambda^*|_1 + \varepsilon_n} \tag{11}$$

where we take $\varepsilon_n = \phi(0)/(nr_n)$.

Our first result states the oracle inequality. It holds true as long as the tuning parameter $r_n$ in the penalty term exceeds $\widehat{r}$.

**Theorem 1.** *Assume that (7) and (10) hold. On the event $r_n > \widehat{r}$,*

$$\Delta_\phi(\widehat{\lambda}) + r_n|\widehat{\lambda} - \lambda^*|_1 \leq 3\Delta_\phi(\lambda^*) + 2\left(\frac{8C_{\Delta,\mu}}{c_\mu}\right)^{\frac{1}{1-\beta}} (r_n^2|\lambda^*|_0)^{\frac{1}{2-2\beta}} + \frac{2\phi(0)}{n}. \tag{12}$$

The next section discusses choices of the tuning parameter $r_n$ that ensure that the probability of the event $\{r_n \geq \widehat{r}\}$ is large.

### 2.4. Choice of the tuning parameter $r_n$

The next lemma states that $\widehat{r}$ is sharply concentrated around its mean.

**Lemma 2.** *Let $C_F = \max_{1 \leq j \leq M} \|f_j\|_\infty$. We have*

$$0 \leq \widehat{r} \leq 2C_\phi C_F \tag{13}$$

*and, for all $\delta > 0$,*

$$\mathbb{P}\{\widehat{r} - \mathbb{E}[\widehat{r}] \geq \delta\} \leq \exp\left(-\frac{1}{2}\frac{n\delta^2}{C_\phi^2 C_F^2}\right) \tag{14}$$

*Proof.* The first assertion follows directly from the definition of $\widehat{r}$. The second statement follows from an application of McDiarmid's bounded differences inequality [10, Theorem 2.2, page 8] after observing that a change of a single pair $(X_i, Y_i)$ changes $\widehat{r}$ by at most $2C_\phi C_F/n$. □

The range of $\widehat{r}$ in (13) is important for implementation of the method: We suggest to find a good value for $r_n$ based on cross validation and the grid can be taken on the interval $[0, 2C_\phi C_F]$. Inequality (14) is important for theoretical considerations. It shows that we should take

$$r_n = \mathbb{E}[\widehat{r}] + \sqrt{\frac{2\log(1/\delta)}{n}} C_\phi C_F \qquad (15)$$

for some $0 < \delta < 1$, since then

$$\mathbb{P}\{r_n \geq \widehat{r}\} \geq 1 - \delta.$$

The expected value $\mathbb{E}[\widehat{r}]$ is of order $\{\log(M \vee n)/n\}^{1/2}$ by the following lemma.

**Lemma 3.** *Let $J_n$ be the smallest integer such that $2^{J_n} \geq n$. Then, for all $M, n \geq 1$ and $0 < \delta < 1$*

$$\mathbb{E}[\widehat{r}] \leq \frac{7C_\phi C_F}{\sqrt{n}}\sqrt{2\log 2(M \vee n)} + \frac{J_n C_\phi C_F}{2(M \vee n)^2}.$$

Consequently,

**Corollary 4.** *Assume that (7) and (10) hold, and take*

$$r_n \geq \frac{7C_\phi C_F}{\sqrt{n}}\sqrt{2\log 2(M \vee n)} + \frac{J_n C_\phi C_F}{2(M \vee n)^2} + C_\phi C_F \sqrt{\frac{2\log(1/\delta)}{n}}. \qquad (16)$$

*Then oracle inequality (12) holds with probability at least $1 - \delta$.*

## 3. Example: generalized hinge loss

Throughout this section, we consider a fixed cost $d$ and a fixed threshold value $\tau$ with $0 \leq d \leq 1/2$ and $d \leq \tau \leq 1 - d$. Instead of the discontinuous loss $\ell(z)$ defined in (3), [1] considers the convex surrogate loss

$$\phi_d(z) = \begin{cases} 1 - az & \text{if } z < 0, \\ 1 - z & \text{if } 0 \leq z < 1, \\ 0 & \text{otherwise} \end{cases} \qquad (17)$$

where $a = (1-d)/d \geq 1$ and shows that the Bayes discriminant function $f_0$ defined in (4) minimizes both the risks $\mathbb{E}[\ell(Yf(X))]$ and $\mathbb{E}[\phi_d(Yf(X))]$ over all measurable $f: \mathcal{X} \to \mathbb{R}$. We see that $\phi_d(z) \geq \ell(z)$ for all $z \in \mathbb{R}$ as long as $0 \leq \tau \leq 1-d$. Moreover, [1] shows that a relation like this holds not only for the loss functions and hence the risks, but for the excess risks as well. In particular, for all $d \leq \tau \leq 1-d$, we have

$$\mathbb{E}\left[\ell(Yf(X))\right] - \mathbb{E}\left[\ell(Yf_0(X))\right] \leq \mathbb{E}\left[\phi_d(Yf(X))\right] - \mathbb{E}\left[\phi_d(Yf_0(X))\right]. \qquad (18)$$

This is important since minimization of (5) produces oracle inequalities in terms of the $\phi_d$-excess risk (Theorem 1), not in terms of the original excess risk directly. The latter risk has a sound statistical interpretation.

For plug-in rules and empirical risk minimizers, [1, 11] show that for classification with a reject option, fast rates (faster than $n^{-1/2}$) for the excess risk may be obtained if the probability that $\eta(X)$, defined in (1), is close to the critical values of $d$ and $1-d$, is small. More precisely, assume that there exist $A \geq 1$ and $\alpha \geq 0$ such that for all $t > 0$,

$$\mathbb{P}\{|\eta(X) - d| \leq t\} \leq At^\alpha \text{ and } \mathbb{P}\{|\eta(X) - (1-d)| \leq t\} \leq At^\alpha. \tag{19}$$

For $d = 1/2$, this asumption is equivalent to Tsybakov's margin condition [15]. Then, [1, Proof of Lemma 7] shows that

$$\Delta_{\phi_d}(\lambda) \geq \frac{\{\mathbb{E}\left[\rho_\eta(\mathsf{f}_\lambda(X), f_0(X))\right]\}^{\frac{1+\alpha}{\alpha}}}{2d\{4A(1+|\lambda|_1 C_F)\}^{\frac{1}{\alpha}}} \tag{20}$$

where

$$\rho_\eta(f, f_0) = \begin{cases} \eta|f - f_0| & \text{if } \eta < d \text{ and } f < -1, \\ (1-\eta)|f - f_0| & \text{if } \eta > 1-d \text{ and } f > 1, \\ |f - f_0| & \text{otherwise.} \end{cases}$$

Following [14], we consider the measure $\mu$ defined by

$$\mu(B) = \int_B \eta(x)\{1 - \eta(x)\}P(dx), \tag{21}$$

for any Borel set $B$, where $P$ is the probability measure of $X$. Since

$$\int \{\mathsf{f}_\lambda(x) - f_0(x)\}^2 \mu(dx) \leq (1 + |\lambda|_1 C_F) \int |\mathsf{f}_\lambda(x) - f_0(x)| \mu(dx),$$

it follows from (20) that condition (7) holds for all $\lambda$ with $|\lambda|_1 \leq C_\Lambda$ with

$$C_{\Delta,\mu} = (1 + C_\Lambda C_F)^{\frac{1+\alpha}{2+2\alpha}} (2d)^{\frac{\alpha}{2+2\alpha}} \{4A(1 + C_\Lambda C_F)\}^{\frac{1}{2+2\alpha}}, \tag{22}$$

and $\beta = \alpha/(2 + 2\alpha)$.

Let $\widehat{\lambda}$ minimize the penalized empirical risk $\widehat{R}_{\phi_d}(\lambda) + p(\lambda)$ over the restricted set

$$\Lambda = \{\lambda \in \mathbb{R}^M : |\lambda|_1 \leq C_\Lambda\}$$

for some finite $C_\Lambda$ and let $\lambda^*$ minimize

$$3\Delta_{\phi_d}(\lambda) + 2\left(\frac{8C_{\Delta,\mu}}{c_\mu}\right)^{\frac{2+2\alpha}{2+\alpha}} \left(r_n^2 |\lambda|_0\right)^{\frac{1+\alpha}{2+\alpha}} \tag{23}$$

over $\lambda \in \Lambda$. Provided then that the mutual coherence assumption (10) holds, Corollary 4 states that for all choices $r_n = r_n(\delta)$ in (16) with $C_\phi = (1-d)/d$,

$$\Delta_{\phi_d}(\widehat{\lambda}) + r_n|\widehat{\lambda} - \lambda^*|_1 \leq 3\Delta_\phi(\lambda^*) + 2\left(\frac{8C_{\Delta,\mu}}{c_\mu}\right)^{\frac{2+2\alpha}{2+\alpha}} (r_n^2|\lambda^*|_0)^{\frac{1+\alpha}{2+\alpha}} + \frac{2\phi(0)}{n}.$$

with probability at least $1-\delta$, where $0 < \delta < 1$ is given in (16). Consequently, via (18),

**Theorem 5.** *Assume that (19) holds for some $\alpha \geq 0$ and that the dictionary $F_M$ satisfies (10) with $\mu$ defined in (21). Let $\lambda^* \in \Lambda$ be as given in (23). Then the minimizer $\widehat{\lambda} \in \Lambda$ with $r_n$ as in (16) with $\delta = 1/(n \vee M)$ and $C_\phi = (1-d)/d$ satisfies, for $C_{\Delta,\mu}$ defined in (22),*

$$\mathbb{E}[\ell(Y\mathsf{f}_{\widehat{\lambda}}(X))] - \mathbb{E}[\ell(Yf_0(X))] + r_n|\widehat{\lambda} - \lambda^*|_1 \leq$$

$$3\Delta_\phi(\lambda^*) + 2\left(\frac{8C_{\Delta,\mu}}{c_\mu}\right)^{\frac{2+2\alpha}{2+\alpha}} (r_n^2|\lambda^*|_0)^{\frac{1+\alpha}{2+\alpha}} + \frac{2\phi(0)}{n}$$

*with probability tending to 1 as $n \to \infty$.*

The best possible "rate" $(r_n^2|\lambda^*|_0)^{(1+\alpha)/(2+\alpha)}$ is achieved at $\alpha = +\infty$. The slowest possible rate is achieved at $\alpha = 0$ in which case (19) imposes no restriction at all on $\eta(X)$.

## 4. Proofs

### 4.1. Proof of Theorem 1

**Lemma 6.** *On the set $\widehat{r} \leq r_n$, we have*

$$\Delta_\phi(\widehat{\lambda}) - \Delta_\phi(\lambda^*) + r_n|\widehat{\lambda} - \lambda^*|_1 \leq 4r_n \sum_{i \in I^*} |\widehat{\lambda}_i - \lambda_i^*| + r_n\varepsilon_n. \qquad (24)$$

*Proof.* Rewrite (6) to obtain, for $\widehat{G}(\lambda) = \widehat{R}(\lambda) - R(\lambda)$,

$$R_\phi(\widehat{\lambda}) - R_\phi(\lambda^*) \leq \widehat{G}(\lambda^*) - \widehat{G}(\widehat{\lambda}) + p(\lambda^*) - p(\widehat{\lambda})$$
$$\leq \widehat{r}|\widehat{\lambda} - \lambda^*|_1 + \varepsilon_n\widehat{r} + p(\lambda^*) - p(\widehat{\lambda}).$$

On the event $r_n \geq \widehat{r}$ then,

$$\Delta_\phi(\widehat{\lambda}) - \Delta_\phi(\lambda^*) \leq r_n|\widehat{\lambda} - \lambda^*|_1 + \varepsilon_n r_n + p(\lambda^*) - p(\widehat{\lambda}).$$

Add $r_n|\widehat{\lambda} - \lambda^*|_1$ to both sides, and deduce

$$\Delta_\phi(\widehat{\lambda}) - \Delta_\phi(\lambda^*) + r_n|\widehat{\lambda} - \lambda^*|_1$$
$$\leq 2r_n|\widehat{\lambda} - \lambda^*|_1 + r_n\varepsilon_n + 2r_n|\lambda^*|_1 - 2r_n|\widehat{\lambda}|_1$$
$$\leq 2r_n \sum_{i \in I^*} |\widehat{\lambda}_i - \lambda_i^*| + 2r_n \sum_{i \notin I^*} |\widehat{\lambda}_i| - 2r_n \sum_{i=1}^M |\widehat{\lambda}_i| + 2r_n \sum_{i \in I^*} |\lambda_i^*| + r_n\varepsilon_n$$
$$\leq 4r_n \sum_{i \in I^*} |\widehat{\lambda}_i - \lambda_i^*| + r_n\varepsilon_n,$$

which proves our claim. □

**Lemma 7.**

$$c_\mu \sum_{i \in I^*} |\widehat{\lambda}_i - \lambda_i^*| \leq 2\rho^* |\widehat{\lambda} - \lambda^*|_1 + |\lambda^*|_0^{1/2} \|f_{\widehat{\lambda} - \lambda^*}\| \quad (25)$$

*Proof.* See the proof of Theorem 2 of [7, pages 536, 537]. For completeness, we repeat the argument: Set

$$u_j = \widehat{\lambda}_j - \lambda_j^*, \quad U^* = \sum_{j \in I^*} |u_j| \|f_j\|, \quad U = \sum_{j=1}^M |u_j| \|f_j\|.$$

Clearly

$$\sum_{i,j \notin I^*} \sum <f_i, f_j> u_i u_j \geq 0$$

and so we obtain

$$\begin{aligned}
\sum_{j \in I^*} u_j^2 \|f_j\|^2 &= \|f_{\widehat{\lambda} - \lambda^*}\|^2 - \sum_{i,j \notin I^*} \sum u_i u_j <f_i, f_j> - 2 \sum_{i \notin I^*} \sum_{j \in I^*} u_i u_j <f_i, f_j> \\
&\quad - \sum_{i,j \in I^*, i \neq j} u_i u_j <f_i, f_j> \\
&\leq \|f_{\widehat{\lambda} - \lambda^*}\|^2 + 2\rho^* \sum_{i \notin I^*} |u_i| \|f_i\| \sum_{j \in I^*} |u_j| \|f_j\| \\
&\quad + \rho^* \sum_{i,j \in I^*} |u_i| |u_j| \|f_i\| \|f_j\| \\
&= \|f_{\widehat{\lambda} - \lambda^*}\|^2 + 2\rho^* U^* U - \rho^* (U^*)^2. \quad (26)
\end{aligned}$$

The left-hand side can be bounded by $\sum_{j \in I^*} u_j^2 \|f_j\|^2 \geq (U^*)^2 / |\lambda^*|_0$ using the Cauchy-Schwarz inequality, and we obtain that

$$(U^*)^2 \leq \|f_{\widehat{\lambda} - \lambda^*}\|^2 |\lambda^*|_0 + 2\rho^* |\lambda^*|_0 U^* U$$

and, using the properties of a function of degree two in $U(\lambda)$, we further obtain

$$U^* \leq 2\rho^* |\lambda^*|_0 U + \sqrt{|\lambda^*|_0} \|f_{\widehat{\lambda} - \lambda^*}\| \quad (27)$$

and the results follows from $c_\mu \sum_{i \in I^*} |\widehat{\lambda}_i^* - \lambda_i^*| \leq U^*$. □

Combining both lemmas with the mutual coherence assumption immediately gives

**Lemma 8.** *On the event $r_n \geq \widehat{r}$,*

$$\Delta_\phi(\widehat{\lambda}) - \Delta_\phi(\lambda^*) + \frac{1}{2} r_n |\widehat{\lambda} - \lambda^*|_1 \leq \frac{4}{c_\mu} r_n |\lambda^*|_0^{1/2} \|f_{\widehat{\lambda} - \lambda^*}\| + r_n \varepsilon_n \quad (28)$$

Finally we use the link between the $L_2(\mu)$ norm of $f_\lambda - f_0$ and the excess risk $\Delta_\phi(\lambda)$ and Young's inequality that states

$$ab \leq \frac{a^p}{p} + \frac{b^q}{q}, \quad p > 1, \ q = \frac{p}{p-1}$$

so that,

$$ab \leq \frac{\delta}{p}a^p + \frac{p-1}{p\delta^{1/(p-1)}}b^{p/(p-1)}$$

for all $a, b, \delta > 0$. From Lemma 3 above and condition (7), on the event $r_n \geq \widehat{r}$,

$$\Delta_\phi(\widehat{\lambda}) - \Delta_\phi(\lambda^*) + \frac{1}{2}r_n|\widehat{\lambda} - \lambda^*|_1 \leq \frac{4C_{\Delta,\mu}}{c_\mu}r_n|\lambda^*|_0^{1/2}\{\Delta_\phi^\beta(\widehat{\lambda}) + \Delta_\phi^\beta(\lambda^*)\} + r_n\varepsilon_n$$

Now use the above Young's inequality twice with $p = 1/\beta$, $\delta = 1/2$, $b = 4|r_n^2\lambda^*|_0^{1/2}C_{\Delta,\mu}/c_\mu$ and $a = \Delta_\phi^\beta(\widehat{\lambda})$ and $a = \Delta_\phi^\beta(\lambda^*)$, respectively, to deduce

$$\Delta_\phi(\widehat{\lambda}) - \Delta_\phi(\lambda^*) + \frac{1}{2}r_n|\widehat{\lambda} - \lambda^*|_1$$

$$\leq \frac{\beta}{2}\left\{\Delta_\phi(\widehat{\lambda}) + \Delta_\phi(\lambda^*)\right\} + (1-\beta)|r_n^2\lambda^*|_0^{\frac{1}{2(1-\beta)}}\left(\frac{8C_{\Delta,\mu}}{c_\mu}\right)^{\frac{1}{1-\beta}} + r_n\varepsilon_n$$

$$\leq \frac{1}{2}\left\{\Delta_\phi(\widehat{\lambda}) + \Delta_\phi(\lambda^*)\right\} + |r_n^2\lambda^*|_0^{\frac{1}{2(1-\beta)}}\left(\frac{8C_{\Delta,\mu}}{c_\mu}\right)^{\frac{1}{1-\beta}} + r_n\varepsilon_n$$

This concludes the proof of Theorem 1. □

### 4.2. Proof of Lemma 3

Let $\sigma_1, \ldots, \sigma_n$ be independent Rademacher variables, taking the values $\pm 1$, each with probability $1/2$, independent of the data $(X_1, Y_1), \ldots, (X_n, Y_n)$. Set

$$\widehat{G}^0(\lambda) = \frac{1}{n}\sum_{i=1}^n \sigma_i\{\phi(Y_i f_\lambda(X_i)) - \phi(Y_i f_{\lambda^*}(X_i))\}$$

A standard symmetrization trick ([10, page 18]) shows that

$$\mathbb{E}[\widehat{r}] \leq \mathbb{E}\left[\sup_{\lambda \in \Lambda_n} \frac{|\widehat{G}^0(\lambda) - \widehat{G}^0(\lambda^*)|}{|\lambda - \lambda^*|_1 + \varepsilon_n}\right]$$

$$\leq \mathbb{E}\left[\sup_{|\lambda - \lambda^*|_1 \leq \varepsilon_n} \frac{|\widehat{G}^0(\lambda) - \widehat{G}^0(\lambda^*)|}{|\lambda - \lambda^*|_1 + \varepsilon_n}\right] + \mathbb{E}\left[\sup_{\varepsilon_n \leq |\lambda - \lambda^*|_1 \leq \phi(0)/r_n} \frac{|\widehat{G}^0(\lambda) - \widehat{G}^0(\lambda^*)|}{|\lambda - \lambda^*|_1 + \varepsilon_n}\right]$$

$$= (I) + (II)$$

as $|\lambda - \lambda^*|_1 \leq \phi(0)/r_n$ for all $\lambda \in \Lambda_n$. The first term can be bounded as follows:

$$
\begin{aligned}
(I) &\leq \frac{1}{\varepsilon_n}\mathbb{E}\left[\sup_{|\lambda-\lambda^*|_1\leq\varepsilon_n}\left|\widehat{G}^0(\lambda) - \widehat{G}^0(\lambda^*)\right|\right] \\
&\leq \frac{C_\phi}{\varepsilon_n}\mathbb{E}\left[\sup_{|\lambda-\lambda^*|_1\leq\varepsilon_n}\left|\frac{1}{n}\sum_{i=1}^n \sigma_i Y_i \mathsf{f}_{\lambda-\lambda^*}(X_i)\right|\right]
\end{aligned}
$$

by the contraction principle for Rademacher processes, see [12, pages 112 – 113]. This implies that

$$
\begin{aligned}
(I) &\leq \frac{C_\phi}{\varepsilon_n}\mathbb{E}\left[\sup_{|\lambda-\lambda^*|_1\leq\varepsilon_n}|\lambda-\lambda^*|_1 \max_{1\leq j\leq M}\left|\frac{1}{n}\sum_{i=1}^n \sigma_i Y_i f_j(X_i)\right|\right] \\
&\leq C_\phi \mathbb{E}\left[\max_{1\leq j\leq M}\left|\frac{1}{n}\sum_{i=1}^n \sigma_i Y_i f_j(X_i)\right|\right] \\
&\leq C_\phi C_F \frac{\sqrt{2\log(2M)}}{\sqrt{n}}
\end{aligned}
$$

where we used [10, Lemma 2.2, page 7] to get the last inequality. We can apply this result since

$$
\mathbb{E}\left[\exp\left\{s\sum_{i=1}^n \sigma_i Y_i f_j(X_i)\right\}\right] \leq \exp(ns^2 C_F^2/2)
$$

for all $s$, that follows in turn from [10, Lemma 2.1, page 5].

The second term (II) requires a peeling argument [16, page 70]. Since $0 \leq \widehat{r} \leq 2C_\phi C_F$ almost surely, we can use the bound

$$
\mathbb{E}[II] \leq \zeta + 2C_\phi C_F \mathbb{P}\{(II) \geq \zeta\}. \tag{29}
$$

Observe that for any $\zeta > 0$, and for $J_n$ the smallest integer with $2^{J_n}\varepsilon_n \geq \phi(0)/r_n$ or $2^{J_n} \geq n$,

$$
\begin{aligned}
&\mathbb{P}\left\{\sup_{\varepsilon_n\leq|\lambda-\lambda^*|_1\leq\phi(0)/r_n}\frac{|\widehat{G}^0(\lambda) - \widehat{G}^0(\lambda^*)|}{|\lambda-\lambda^*|_1+\varepsilon_n} \geq \zeta\right\} \\
&\leq \sum_{j=1}^{J_n}\mathbb{P}\left\{\sup_{2^{j-1}\varepsilon_n\leq|\lambda-\lambda^*|_1\leq 2^j\varepsilon_n}\frac{|\widehat{G}^0(\lambda) - \widehat{G}^0(\lambda^*)|}{|\lambda-\lambda^*|_1+\varepsilon_n} \geq \zeta\right\} \\
&\leq \sum_{j=1}^{J_n}\mathbb{P}\left\{\sup_{2^{j-1}\varepsilon_n\leq|\lambda-\lambda^*|_1\leq 2^j\varepsilon_n}\left|\widehat{G}^0(\lambda) - \widehat{G}^0(\lambda^*)\right| \geq 2^{j-1}\varepsilon_n\zeta\right\}
\end{aligned}
$$

Now, set

$$
Z_j = \sup_{|\lambda-\lambda^*|_1\leq 2^j\varepsilon_n}\left|\widehat{G}^0(\lambda) - \widehat{G}^0(\lambda^*)\right|
$$

and the same considerations leading to the final bound of (I) above yield

$$\mathbb{E}[Z_j] \leq 2^j \varepsilon_n C_\phi C_F \frac{\sqrt{2\log(2M)}}{\sqrt{n}}$$

and

$$\sum_{j=1}^{J_n} \mathbb{P}\left\{\sup_{2^{j-1}\varepsilon_n \leq |\lambda-\lambda^*|_1 \leq 2^j\varepsilon_n} \left|\widehat{G}^0(\lambda) - \widehat{G}^0(\lambda^*)\right| \geq 2^{j-1}\varepsilon_n\zeta\right\}$$
$$\leq \sum_{j=1}^{J_n} \mathbb{P}\left\{Z_j - \mathbb{E}[Z_j] \geq 2^{j-1}\varepsilon_n\zeta - \mathbb{E}[Z_j]\right\}.$$

A change of a single pair $(X_i, Y_i)$ changes $Z_j$ by at most $2C_\phi C_F(2^j\varepsilon_n)/n$, so that another application of the bounded differences inequality [10, Theorem 2.2, page 8] gives, by taking

$$\zeta = 6C_\phi C_F \frac{\sqrt{2\log 2(M \vee n)}}{\sqrt{n}},$$

the final bound

$$\sum_{j=1}^{J_n} \mathbb{P}\left\{Z_j - \mathbb{E}[Z_j] \geq 2^{j-1}\varepsilon_n\zeta - \mathbb{E}[Z_j]\right\}$$
$$\leq \sum_{j=1}^{J_n} \mathbb{P}\left\{Z_j - \mathbb{E}[Z_j] \geq 2 \cdot 2^j\varepsilon_n \frac{\sqrt{2\log(2M \vee 2n)}}{\sqrt{n}}\right\}$$
$$\leq J_n \exp\left\{-\frac{2(C_\phi C_F 2^j\varepsilon_n)^2 2\log(2M \vee 2n)}{(C_\phi C_F 2^j\varepsilon_n)^2}\right\}$$
$$= J_n \exp\left\{-2\log(2M \vee 2n)\right\}$$
$$= J_n(2M \vee 2n)^{-2}.$$

Invoke (29) to conclude the proof of Lemma 3. □

*M. Wegkamp/Lasso type classifiers with a reject option* 168[4] F. Bunea, A.B. Tsybakov and M.H. Wegkamp (2004). Aggregation for Gaussian regression. Technical Report M972, Department of Statistics, Florida State University. To appear in the *Annals of Statistics*.

[5] F. Bunea, A.B. Tsybakov and M.H. Wegkamp (2006). Aggregation and sparsity via $\ell_1$-penalized least squares. *Proceedings of 19th Annual Conference on Learning Theory, COLT 2006. Lecture Notes in Artificial Intelligence*, 4005, 379–391. Springer-Verlag, Heidelberg. MR2280619

[6] F. Bunea, A.B. Tsybakov and M.H. Wegkamp (2006). Sparsity oracle inequalities for the Lasso. Technical Report M979, Department of Statistics, Florida State University.

[7] F. Bunea, A.B. Tsybakov and M.H. Wegkamp (2007). Sparse density estimation with $\ell_1$ penalties. *Proceedings of 20th Annual Conference on Learning Theory, COLT 2007. Lecture Notes in Artificial Intelligence*, 4539, 530–543. Springer-Verlag, Heidelberg.

[8] F. Bunea and M.H. Wegkamp (2004). Two-stage model selection procedures in partially linear regression. *Canadian Journal of Statistics*, 32(2), 105–118. MR2064395

[9] C.K. Chow (1970). On optimum error and reject trade-off. *IEEE Transactions on Information Theory*, 16, 41–46.

[10] L. Devroye and G. Lugosi (2000). *Combinatorial Methods in density estimation*. Springer, New York. MR1843146

[11] R. Herbei and M. H. Wegkamp (2006). Classification with reject option. *Canadian Journal of Statistics*, 34(4), 709–721.

[12] M. Ledoux and M. Talagrand (1991). *Probability in Banach Spaces*. Springer, New York. MR1102015

[13] B. D. Ripley (1996). *Pattern recognition and neural networks*. Cambridge University Press, Cambridge. MR1438788

[14] B. Tarigan and S. A. van de Geer (2006). Classifiers of support vector machine type with $\ell_1$ complexity regularization. *Bernoulli*, 12(6), 1045–1076 MR2274857

[15] A. B. Tsybakov (2004). Optimal aggregation of classifiers in statistical learning. *Annals of Statistics*, 32, 135–166. MR2051002

[16] S.A. van de Geer (2000). *Empirical processes in M-estimation*. Cambridge Series in statistical and Probabilistic Mathematics. Cambridge University Press, Cambridge.

[17] S.A. van de Geer (2006). High dimensional generalized linear models and the Lasso. Research report No.133. Seminar für Statistik, ETH.